\documentclass[10pt,twocolumn,letterpaper]{article}

\usepackage{cvpr}
\usepackage{times}
\usepackage{epsfig}
\usepackage{graphicx}
\usepackage{amsmath}
\usepackage{amssymb}
\usepackage{bm}
\usepackage{soul}
\usepackage[dvipsnames]{xcolor}
\usepackage[linesnumbered,ruled,lined,boxed]{algorithm2e}

\usepackage[breaklinks=true,bookmarks=false]{hyperref}

\cvprfinalcopy % *** Uncomment this line for the final submission

\def\cvprPaperID{22} % *** Enter the CVPR Paper ID here

\def\eva{{a}}
\def\evx{{x}}
\def\rx{{\textnormal{x}}}
\def\va{{\bm{a}}}
\def\vw{{\bm{w}}}
\def\vr{{\bm{r}}}
\def\vu{{\bm{u}}}
\def\vv{{\bm{v}}}
\def\vx{{\bm{x}}}
\def\vs{{\bm{s}}}
\def\mX{{\bm{X}}}
\def\mS{{\bm{S}}}
\newcommand{\R}{\mathbb{R}}
\newcommand{\E}{\mathbb{E}}
\newcommand{\qed}{\blacksquare}

\begin{document}

%%%%%%%%% TITLE
\title{Least squares binary quantization of neural networks}

\author{Hadi Pouransari, Zhucheng Tu, Oncel Tuzel\\
Apple Inc.\\
Cupertino, CA 95014, USA \\
%\texttt{\{mpouransari,zhucheng_tu,ctuzel\}@apple.com} \\
}

\maketitle

%-------------------------------------------------------------------------
%%%%%%%%% ABSTRACT
%-------------------------------------------------------------------------
\begin{abstract}
Quantizing weights and activations of deep neural networks results in significant improvement in inference efficiency at the cost of lower accuracy. A source of the accuracy gap between full precision and quantized models is the quantization error.
In this work, we focus on the binary quantization, in which values are mapped to -1 and 1. 
We provide a unified framework to analyze different scaling strategies.
Inspired by the pareto-optimality of 2-bits versus 1-bit quantization, we introduce a novel 2-bits quantization with provably least squares error.
Our quantization algorithms can be implemented efficiently on the hardware using bitwise operations.
We present proofs to show that our proposed methods are optimal, and also provide empirical error analysis.
We conduct experiments on the ImageNet dataset and show a reduced accuracy gap when using the proposed least squares quantization algorithms.\footnote{\url{https://github.com/apple/ml-quant}}
\end{abstract}

%-------------------------------------------------------------------------
%%%%%%%%% INTRODUCTION
%-------------------------------------------------------------------------
\section{Introduction}

A major challenge in the deployment of Deep Neural Networks (DNNs) is their high computational cost. 
Finding effective methods to improve run-time efficiency is still an area of research. We can group various approaches taken by researchers into the following three categories.

{\bf Hardware optimization}: Specifically designed hardwares are deployed to efficiently perform computations in ML tasks.
{\bf Compiler optimization}: Compression and fusion techniques coupled with efficient hardware-aware implementations, such as dense and sparse matrix-vector multiplication, are used.
{\bf Model optimization}: 
Run-time performance can also be gained by modifying the model structure and the underlying arithmetic operations. While hardware and compiler optimization are typically lossless, model optimization trades-off computational cost (memory, runtime, or power) for model accuracy. For example, by scaling the width of the network~\cite{zagoruyko2016wide}. The goal of model optimization is to improve the trade-off between computational cost and model accuracy. This work falls into this category. We briefly explain different model optimization techniques here.

%-------------------------------------------------------------------------
{\bf Architecture optimization}
One strategy to construct efficient DNNs is to define a template from which efficient computational blocks can be generated. SqueezeNet~\cite{iandola2016squeezenet}, MobileNets~\cite{howard2017mobilenets,sandler2018mobilenetv2},  ShuffleNets~\cite{ma2018shufflenet,zhang2018shufflenet}, and ESPNets \cite{mehta2018espnet,mehta2019espnetv2} fall into this category. 
Complementary to these methods, NASNet~\cite{zoph2018learning} and EfficientNet~\cite{tan2019efficientnet} search for an  optimal composition of blocks restricted to a computational budget (e.g., FLOPS) by changing the resolution, depth, width, or other parameters of each layer.

%-------------------------------------------------------------------------
{\bf Pruning and Compression}:
Several methods have been proposed to improve runtime performance by detecting and removing computational redundancies. Examples of methods in this category include low-rank acceleration~\cite{jaderberg2014speeding}, the use of depth-wise convolution in Inception~\cite{szegedy2015going}, sparsification of kernels in deep compression~\cite{han2015deep}, re-training redundant neurons in DSD~\cite{han2016dsd}, depth-wise separable convolution in Xception~\cite{chollet2017xception}, pruning redundant filters in PFA~\cite{suau2018network}, finding an optimal sub-network in lottery ticket hypothesis~\cite{frankle2018lottery}, and separating channels based on the features resolution in octave convolution~\cite{chen2019drop}.

%-------------------------------------------------------------------------
{\bf Low-precision arithmetic and quantization:}
Another avenue to improve runtime performance, and the focus of this work, is to use low-precision arithmetic. The idea is to use fewer bits to represent weights and activations. Some instances of these strategies already exist in AI compilers, where it is common to cast weights of a trained model from 32 bits to 16 or 8 bits. However, in general, post-training quantization reduces the model accuracy. This can be addressed by incorporating lower-precision arithmetic into the training process (during-training quantization), allowing the resulting model to better adapt to the lower precision~\cite{gupta2015deep,jacob2018quantization}. Additionally, to improve utilization, many works considered mixed-precision quantization, where different number of bits are allowed at different layers of the network~\cite{dong2019hawq,wang2019haq,wu2018mixed}.

Using fewer bits results in dramatic memory savings. This has motivated research into methods that use a single bit to represent a scalar weight: In \cite{courbariaux2015binaryconnect} the authors train models with weights quantized to the values in $\{-1, 1\}$. While this results in a high level of compression, model accuracy can drop significantly. \cite{li2016ternary} and \cite{zhu2016trained} reduce the accuracy gap between full precision and quantized models by considering ternary quantization (using the values in $\{-1, 0, 1\}$), at the cost of slightly less compression.

\begin{figure}[htb!]
\begin{center}
\includegraphics[scale=.9]{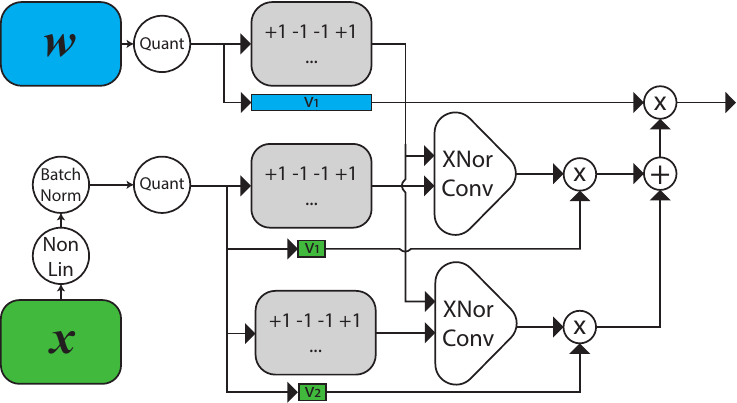}
\end{center}
\caption{When both weights and activations are quantized using binary quantization, the convolution can be implemented efficiently using bitwise XNor and bit-counting operations. See Section \ref{sec:scalar_sign_appx_high_order} for more details.}
\label{fig:arch}
\end{figure}

To further improve the computational efficiency, the intermediate activation tensors (feature maps) can also be quantized. When this is the case, an implementation can use high-performance operators that act on quantized inputs, for example a convolutional block depicted in Figure \ref{fig:arch}. This idea has been explored in~\cite{cai2017deep,courbariaux2016binarized,hubara2017quantized,lin2017towards,mishra2017wrpn,park2018value,rastegari2016xnor,zhang2018lq,zhou2016dorefa}, and many other works.

We call a mapping from a tensor with full precision entries to a tensor with the same shape but with values in $\{-1, 1\}$ a {\bf binary quantization}. When both weights and activations of a DNN are quantized using binary quantization, called Binary Neural Network (BNN), fast and power-efficient kernels which use bitwise operations can be implemented. Observe that the inner-product between two vectors with entries in $\{-1, 1\}$ can be written as bitwise XNor operations followed by bit-counting~\cite{courbariaux2016binarized}. However, the quantization of both weights and activations further reduces the model accuracy~\cite{hubara2017quantized,rastegari2016xnor}.

\begin{figure}[htb!]
\begin{center}
\includegraphics[scale=.19]{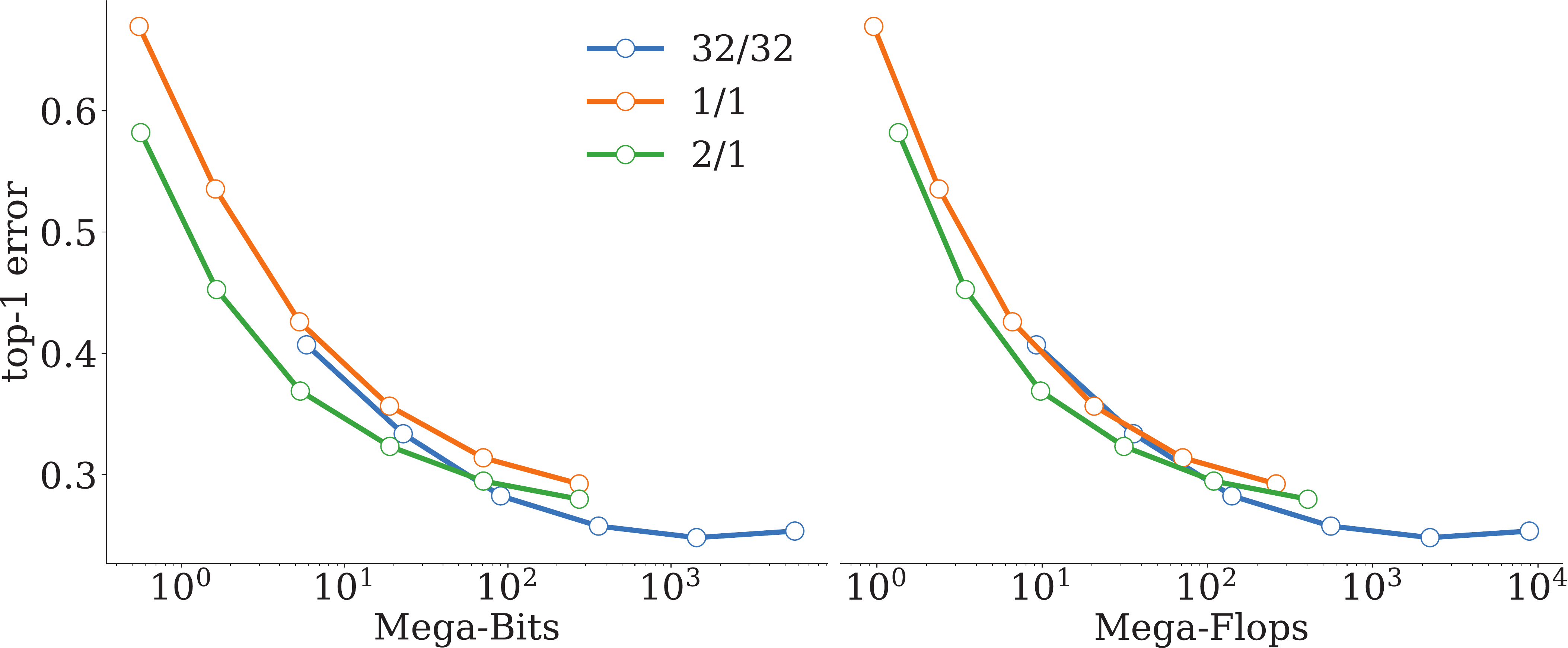}
\end{center}
\caption{The trade-off between computational cost model error for ResNet18 trained on CIFAR100. Left plot show memory foot-print, and right plot shows flop counts. $k^a/k^w$ refers to using $k^a$ bits for activations and $k^w$ bits for weights.}
\label{fig:pareto}
\end{figure}

%-------------------------------------------------------------------------
The accuracy drop due to quantization can be compensated by increasing model capacity through architecture modifications, for example by increasing the number of filters as in~\cite{zagoruyko2016wide}. In Figure \ref{fig:pareto} we show the trade-off between computational cost (memory and flops) and classification error for different quantization schemes by uniformly scaling the number of filters in ResNet18 architecture trained on CIFAR100 dataset~\cite{krizhevsky2009learning}. We followed methodology in \cite{wang2019learning} to approximate equivalent flops of BNNs. We use the same training setup for quantized models without any tuning. This empirical result suggests that for a given computational budget using 2-bits quantizations is pareto-optimal when compared to the original BNN and the full-precision network. \cite{Kim2020BinaryDuo}~also observed much larger accuracy degradation when the bit precision is reduced from 2-bits to 1-bit than other cases with $>\!2$-bits. This interesting observation, although not necessarily a universal conclusion, motivates deriving the optimal 2-bits quantization.

Note that there are many other directions to improve the accuracy of BNNs. For example, using learned clipping in PACT~\cite{choi2018pact}, double skip-connection in BiRealNet~\cite{liu2018bi}, parametric ReLU in~\cite{bulat2019improved}, multi-stage knowledge distillation in~\cite{bagherinezhad2018label,bulat2019improved}, and tailored binary optimization in~\cite{helwegen2019latent}. These are all orthogonal improvements to the proposed provably least squares error 2-bits quantization, and can be used together to further improve the model accuracy.
%-------------------------------------------------------------------------
\subsection{Main contributions}
In this work, we analyze the accuracy of binary quantization when applied to both weights and activations of a DNN, and propose methods to improve the quantization accuracy:
\begin{itemize}
\itemsep0em
\item We present a unified framework to analyze different scaling strategies for BNNs, and show that scaled binary quantization is a good approximation (Section \ref{sec:lr_sign_appx}).
\item We derive 2-bits (Section \ref{sec:2bits_sign_appx}) and ternary (Section \ref{sec:ternary_sign_appx}) scaled binary quantization algorithms with least squares error. We also propose greedy variants of these algorithms (Section \ref{sec:greedy_foldable}).
\item Experiments on the ImageNet dataset show that the optimal algorithms have reduced quantization error, and lead to improved classification accuracy (Section \ref{sec:experiments}).
\end{itemize}

%-------------------------------------------------------------------------
%%%%%%%%% LOW-RANK QUANTIZATION
%-------------------------------------------------------------------------
\section{Low-rank binary quantization}\label{sec:lr_sign_appx}
Binary quantization (that maps entries of a tensor to $\{-1,1\}$) of weights and activation tensors of a neural network can significantly reduce the model accuracy. A remedy to retrieve this accuracy loss is to scale the binarized tensors with few full precision values. For example, \cite{hubara2017quantized} learn a scaling for each channel from the parameters of batch-normalization, and \cite{rastegari2016xnor} scale the quantized activation tensors using the channel-wise average of pixel values.

In this section, using low-rank matrix analysis, we analyze  different scaling strategies. We conclude that multiplying the quantized tensor by a single scalar, which is computationally the most efficient option, has approximately the same accuracy as the more expensive alternatives.

We introduce the rank-1 binary quantization-- an approximation to a matrix $\mX\in \R^{m\times n}$:
\begin{equation}
	\mX \simeq \mX_1 \odot \mS,
\end{equation}

where $\mX_1 \in \R^{m \times n}$ is a rank-$1$ matrix, $\mS \in \{-1,1\}^{m \times n}$, and $\odot$ is element-wise multiplication (Hadamard product). Note that this approximation is also defined for tensors, after appropriate reshaping. For example, for an image classification task, we can reshape the output of a layer of a DNN with shape $h\times w \times n$, where $h$, $w$, and $n$ are height, width, and number of channels, respectively, into an $m \times n$ matrix with $m=hw$ rows and one column per channel.

We define the error of a rank-$1$ binary quantization as $\| \mX - \mX_1 \odot \mS \|_F$, where $\|~\|_F$ is the Frobenius norm. Entries of $\mS$ are in $\{-1,1\}$, therefore, the quantization error is equal to $\| \mX \odot \mS - \mX_1 \|_F$. Note that $\| \mX \odot \mS \|_F^2$ (the total energy), which is equal to sum of the squared singular values, is the same for any $\mS$. Different choices of $\mS$ change the distribution of the total energy among components of the Singular Value Decomposition (SVD) of $\mX \odot \mS$. The optimal rank-$1$ binary quantization is achieved when most of the energy of $\mX \odot \mS$ is in its first component.

In \cite{rastegari2016xnor}, the authors proposed to quantize the activations by applying the sign function and scale them by their channel-wise average. We can formulate this scaling strategy as a special rank-1 binary quantization $\mX \simeq \va \boldsymbol{1}^{\top} \odot ~\text{sign}(\mX)$, where
\begin{equation}
 \begin{aligned}
&\eva_i = \frac{\sum_{j=1}^n |\mX_{i,j}|}{n} \quad \text{for} \quad 1\le i \le m \text{,} \\
&\text{sign}(x) = \begin{cases}
-1 & \text{if $x<0$}\\
1 & \text{if $x \ge 0$}\\
\end{cases},
\end{aligned}
\end{equation}
and $\boldsymbol{1}$ is an $n$-dimensional vector with all entries 1.

In Appendix \ref{sec:opt_rank1} we show that the optimal rank-1 binary quantization of an arbitrary $\mX \in \R^{m \times n}$ is given by $\mS = \text{sign}(\mX)$ and $\mX_1 = \text{truncated}_1\text{-SVD}(|\mX|)$, where $\text{sign}(\mX)$ is the element-wise sign of $\mX$, and $\text{truncated}_1\text{-SVD}(|\mX|) = \sigma_1 \vu_1 \vv_1^{\top}$ is the first component of the SVD of $\mX \odot ~\! \text{sign}(\mX) = | \mX |$.
Moreover, if entries of $\mX$ are i.i.d. $\sim \mathcal{N}(0,1)$, the first singular value of $| \mX |$ captures most of the energy $\sigma_1^2(|\mX|) / \| \mX\|_F^2 \simeq 0.64$, and the first left and right singular vectors are almost constant vectors. Normal distribution assumption for entries of $\mX$ is relevant due to application of Batch Normalization (BN)~\cite{ioffe2015batch}.
Therefore, a scalar multiple of $\text{sign}(\mX)$ approximates $\mX$ well: $\mX \simeq  \sigma_1 \vu_1 \vv_1^{\top} \odot~\! \text{sign}(\mX) \simeq v~\boldsymbol{1} \boldsymbol{1}^{\top} \odot ~\!\text{sign}(\mX) = v~\!\text{sign}(\mX)$, where $v \in \R_{\ge 0}$.
We call this computationally efficient approximation {\bf scaled binary quantization}.

%-------------------------------------------------------------------------
%%%%%%%%% SCALED BINARY QUANTIZATION
%-------------------------------------------------------------------------
\section{Scaled binary quantization} \label{sec:scalar_sign_appx}
In Section \ref{sec:lr_sign_appx} we showed that scaled binary quantization is a good approximation to activation and weight tensors of a DNN.
Next we show how we can further improve the accuracy of scaled binary quantization using more bits.
To simplify the presentation (1) we flatten matrix $\mX \in \R^{m\times n}$ in to a vector $\vx \in \R^N$ with $N=mn$, and
(2) we assume the entries of $\vx$ are different realizations of a random variable $\rx$ with an underlying probability distribution $p(x)$. 
In practice, we compute all statistics using their unbiased estimators from vector $\vx$ (e.g., $\sum_{i} \evx_i / N$ is an unbiased estimator of $\E_{\rx\sim p} [ \rx ]$).
Furthermore, for $f: \R \to \R$, we denote entrywise application of $f$ to $\vx$ by $f(\vx)$.
The quantized approximation of $\vx$ is denoted by $\vx^q$, and the error (loss) of quantization is $\|\vx - \vx^q\|_2$. All optimal solutions refers to the least squares error and hold for an arbitrary distribution $p(x)$.
%------------------------------------------------------------------------- 
\subsection{$1$-Bit quantization}\label{sec:1bit_quant}
A 1-bit scaled binary quantization of $\vx$ is:
\begin{equation} \label{eqn:1bit_sign}
	\vx \simeq \vx^q = v s(\vx),
\end{equation}
which is determined by a scalar $v \in \R_{\ge 0}$ and a function $s: \R \to \{-1,1\}$.
Finding the least squares 1-bit scaled binary quantization can be formulated as the following optimization problem:
\begin{equation}\label{eqn:order1_sign_appx}
    \begin{aligned}
        \underset{v,s}{\text{minimize}} \quad & \int_{-\infty}^{+\infty} p(x)(v~\!s(x) - x)^2 dx\\
        \text{s.t.} \quad & s: \R \to \{-1,1\}, v \in \R_{\ge 0}
    \end{aligned}
\end{equation}

%------------------------------------------------------------------------- 
\subsubsection{Least squares $1$-Bit algorithm}\label{sec:optimal_1bit_alg}
The solution of problem (\ref{eqn:order1_sign_appx}) is given by $v = \E_{\rx\sim p} [ |\rx| ]$ and $s(x) = \text{sign}(x)$ (for the proofs see Appendix \ref{sec:opt_scalar_appx}). Therefore, for a vector $\vx$ the optimal scaled binary quantization is given by
\begin{equation} \label{eqn:opt_order1_sign_appx}
	\vx \simeq \vx^q = \frac{\sum_{i} |\evx_i|}{N} ~\! \text{sign}(\vx),
\end{equation}
where $\frac{\sum_{i} |\evx_i|}{N}$ is an unbiased estimator of $\E_{\rx\sim p} [ |\rx| ]$.

%------------------------------------------------------------------------- 
\subsection{$k$-Bits quantization} \label{sec:scalar_sign_appx_high_order}
We can further improve the accuracy of scaled binary quantization by adding more terms to the approximation (\ref{eqn:1bit_sign}).
A $k$-bits scaled binary quantization of $\vx$ is 
\begin{equation}\label{eqn:kbit_representable}
	\vx \simeq \vx^q = \sum_{i=1}^k v_i s_i(\vx),
\end{equation}
which is determined by a set of  $k$ pairs of scalars $v_i$'s and functions $s_i: \R \to \{-1,1\}$. Observe that any permutation of $(v_i, s_i)$'s results in the same quantization. To remove ambiguity, we assume $v_1 \ge \ldots \ge v_k \ge 0$.

When both weights, $\vw$, and activations, $\vx$, are quantized using scaled binary quantization (\ref{eqn:kbit_representable}),
their inner-product can be written as:
\begin{equation} \label{eqn:xnor_dot}
	\langle \vx^q,  \vw^q \rangle = \sum_{i=1}^{k^a} \sum_{j=1}^{k^w} v_i^a v_j^w \langle \vs_i^a, \vs_j^w \rangle,
\end{equation}
where $ \vx^q = \sum_{i=1}^{k^a} v_i^a \vs_i^a $ and $\vw^q = \sum_{j=1}^{k^w} v_j^w \vs_i^w$ are quantized activations and weights  with $k^a$ and $k^w$ bits, respectively, 
$\vs_i^a = s_i^a(\vx)$, and $\vs_j^w = s_j^w(\vw)$. This inner-product can be computed efficiently using bitwise XNors followed by bit-counting (see Figure \ref{fig:arch} with $k^a = 2$ and $k^w=1$).

Finding the least squares $k$-bits scaled binary quantization can be formulated as:
\begin{equation}\label{eqn:opt_kbit_scalar_appx}
\begin{aligned}
\underset{s_i, v_i}{\text{minimize}} \quad & \int_{-\infty}^{+\infty} p(x)\left(\left(\sum_{i=1}^k v_i s_i(x) \right) - x\right)^2 dx\\
\text{s.t.} \quad & \forall~1\le i \le k \quad s_i: \R \to \{-1,1\},\\ &v_1 \ge v_2 \ge \ldots \ge v_k \ge 0
\end{aligned}
\end{equation}
This is an optimization problem with a non-convex domain for all $k \ge 1$. We solve the optimization for $k=1$ in Section \ref{sec:1bit_quant} and $k=2$ in Section \ref{sec:2bits_sign_appx} for arbitrary distribution $p(x)$. We also provide an approximate solution to (\ref{eqn:opt_kbit_scalar_appx}) in Section \ref{sec:greedy_foldable} using a greedy algorithm.

{\bf Discussion:} A general $k$-bits quantizer maps full precision values to an arbitrary set of $2^k$ numbers, not necessarily in the form of (\ref{eqn:kbit_representable}). The optimal quantization in this case can be computed using the Lloyd's algorithm~\cite{lloyd1982least}. While a general $k$-bits quantization has more representation power compared to $k$-bits scaled binary quantization, it does not allow an efficient implementation based on bitwise operations. Fixed-point representation (as opposed to floating point) is also in the form of (\ref{eqn:kbit_representable}) with an additional constant term. 
However, fixed-point quantization uniformly quantizes the space, therefore, it can be significantly inaccurate for small values of $k$.

%------------------------------------------------------------------------- 
\subsubsection{Foldable quantization} \label{sec:foldable}
In this section, we introduce a special family of $k$-bits scaled binary quantizations that allow fast computation of the quantized values. We name this family of quantizations {\bf foldable}. A $k$-bits scaled binary quantization given by  $(v_i, s_i)$'s is foldable if the following conditions are satisfied:
\begin{equation} \label{eqn:foldable}
\quad \quad s_i(x) = \text{sign}(x - \sum_{j=1}^{i-1} v_j s_j(x)) \quad  \text{for } 1\le i \le k
\end{equation}
When the foldable condition is satisfied, given $v_i$'s, we can compute the $s_i(\vx)$'s in (\ref{eqn:kbit_representable}) efficiently by applying the sign function.

%------------------------------------------------------------------------- 
\subsubsection{Least squares $2$-bits algorithm} \label{sec:2bits_sign_appx}
\begin{figure}[htb!]
\begin{center}
\includegraphics[scale=0.26]{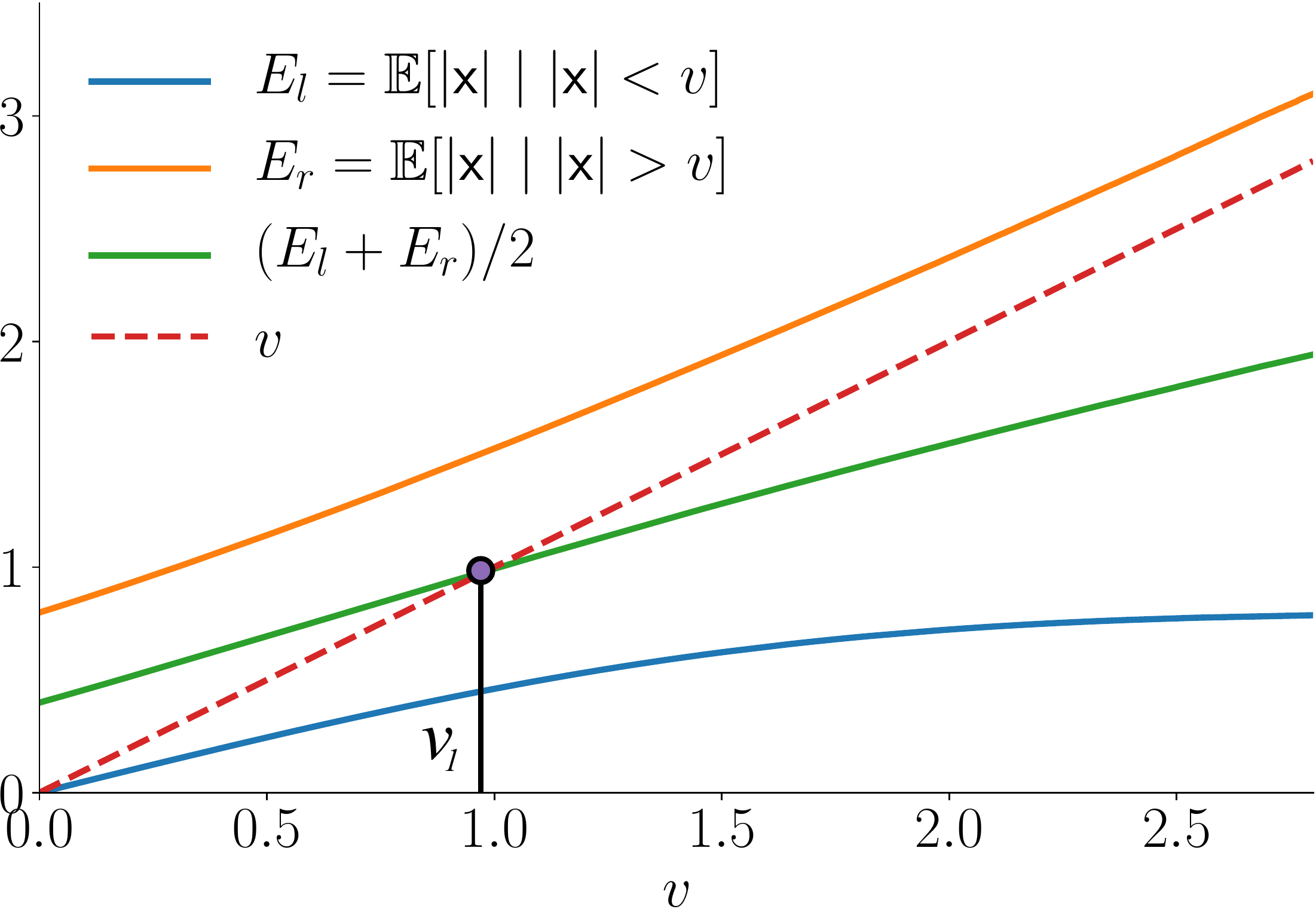}
\end{center}
\caption{The conditional expectations in (\ref{eqn:v1}) for a random variable $\rx$ with standard normal distribution. The optimal value for 2-bits quantization is shown with a solid dot.}
\label{fig:opt_processl}
\end{figure}

In this section, we present the least squares 2-bits binary quantization algorithm, the solution of (\ref{eqn:opt_kbit_scalar_appx}) for $k=2$. In  Appendix \ref{sec:optimal_2bits} we show that the least squares 2-bits binary quantization is foldable and the scalars $v_1$ and $v_2$ should satisfy the following optimality conditions:
\begin{gather}
\label{eqn:v1}
v_1 = \frac{1}{2} \left(  \E_{\rx\sim p} [ |\rx| ~\bm{|}~ |\rx| > v_1] +  \E_{\rx\sim p}[ |\rx| ~\bm{|}~ |\rx| \le v_1] \right)\\
\label{eqn:v2}
v_2 = \frac{1}{2} \left(  \E_{\rx\sim p} [ |\rx| ~\bm{|}~ |\rx| > v_1] -  \E_{\rx\sim p}[ |\rx| ~\bm{|}~ |\rx| \le v_1] \right)
\end{gather}
In Figure \ref{fig:opt_processl} we visualize the conditional expectations that show up in (\ref{eqn:v1}) for a random variable $\rx$ with standard normal distribution. The optimal $v_1$ lies on the intersection of the identity line and average of the conditional expectations in  (\ref{eqn:v1}).

For a given vector $\vx \in \R^N$ we can solve for $v_1$ in (\ref{eqn:v1}) efficiently. We substitute the conditional expectations in (\ref{eqn:v1}) by conditional average operators as their unbiased estimators. (\ref{eqn:v1}) implies that for the optimal $v_1$, the average of the entries in $|\vx|$ smaller than $v_1$ (an estimator of $\E_{\rx\sim p}[ |\rx| ~\bm{|}~ |\rx| \le v_1]$ ) and the average of the entries greater than $v_1$ (an estimator of $\E_{\rx\sim p} [ |\rx| ~\bm{|}~ |\rx| > v_1] $) should be equidistant form $v_1$. 
Note that (\ref{eqn:v1}) may have more than one solution, which are local minima of the objective function in (\ref{eqn:opt_kbit_scalar_appx}).
We find all the values that satisfy this condition in $\mathcal{O}(N \log N)$ time. We first sort entries of $\vx$ based on their absolute value and compute their cumulative sum. Then with one pass we can check whether (\ref{eqn:v1}) is satisfied for each element of $\vx$. We evaluate the objective function in (\ref{eqn:opt_kbit_scalar_appx}) for each local minima, and retain the best.
After $v_1$ is calculated $v_2$ is simply computed from (\ref{eqn:v2}).
As explained in Section \ref{sec:train_and_deploy}, this process is only done during the training. In our experiments, finding the least squares 2-bits quantization was as fast as the 2-bits greedy algorithm (see Section \ref{sec:greedy_foldable}).
Since the least squares 2-bits binary quantization is foldable, after recovering $v_1$ and $v_2$, we have $s_1(\vx) = \text{sign}(\vx)$ and $s_2(\vx) =\text{sign}(\vx - v_1 \text{sign}(\vx))$.

%------------------------------------------------------------------------- 
\subsubsection{Least squares ternary algorithm} \label{sec:ternary_sign_appx}
The boundaries of the optimization domain in (\ref{eqn:opt_kbit_scalar_appx}) for $k=2$, $v_2=0$ and $v_1=v_2=v$, correspond to 1-bit binary and ternary~\cite{li2016ternary} quantizations, respectively.
The scaled ternary quantization maps each full precision value $x$ to $\{-2v, 0 , 2v\}$. Ternary quantization needs 2-bits for representation. However, when a hardware with sparse calculation support is available, for example as in EIE~\cite{han2016eie}, using ternary quantization can be more efficient compared to a general 2-bits quantization.
Setting $v_1=v_2=v$ in (\ref{eqn:v1}) and (\ref{eqn:v2}) we get the ternary optimality condition:
\begin{equation}\label{eqn:vt}
 v =  \frac{1}{2} ~ \E_{\rx\sim p} [ |\rx| ~\bm{|}~ |\rx| > v]
\end{equation}
The process of solving for $v$ in (\ref{eqn:vt}) is similar to that of solving for $v_1$ in (\ref{eqn:v1}) as described above.
The optimality condition in (\ref{eqn:vt}) has been also obtained in~\cite{hou2018loss} using a different approach.

%------------------------------------------------------------------------- 
\subsubsection{$k$-bits greedy algorithm} \label{sec:greedy_foldable}
In this section, we propose a greedy algorithm to compute $k$-bits scaled binary quantization, which we call Greedy Foldable (GF). 
It is given in Algorithm \ref{alg:greedy}.

\LinesNotNumbered
\begin{algorithm}
\caption{$k$-bits Greedy Foldable (GF) binary quantization: compute $\vx^q$ given $\vx$}
\label{alg:greedy}
$\vr \gets \vx$\\
\For{$i\gets1$ \KwTo $k$}{
$v_i$ $\gets$ {mean(abs($\vr$))}\\
$\vs_i$ $\gets$ {sign($\vr$)} \tcp{element-wise sign. For gradient of sign use STE.}
$\vr$ $\gets$ {$\vr - v_i \vs_i$} \tcp{compute new residual.}
    }
\Return {$\vx - \vr$}
\end{algorithm}
In GF algorithm we compute a sequence of residuals. At each step, we greedily find the best $\vs_i$ and $v_i$ for the current residual using the least squares 1-bit binary quantization (\ref{eqn:opt_order1_sign_appx}).
Note that for $k=1$ the GF is the same as the least squares 1-bit binary quantization.

Few of the other papers that have tackled the $k$-bits binary quantization to train quantized DNNs are as follows.
In ReBNet~\cite{ghasemzadeh2018rebnet}, the authors proposed an algorithm similar to Algorithm \ref{alg:greedy}, but considered $v_i$'s as trainable parameters to be learned by back-propagation. \cite{lin2017towards} and \cite{zhang2018lq} find $k$-bits binary quantization via alternating optimization for $s_i$'s and $v_i$'s.
Note that, all these methods produce sub-optimal solutions.

%-------------------------------------------------------------------------
%%%%%%%%% TRAINING
%-------------------------------------------------------------------------
\section{Training binary networks} \label{sec:train_and_deploy}
The loss functions in our quantized neural networks are non-differentiable due to the sign function in the quantizers.
To address this challenge we use the training algorithm proposed in~\cite{courbariaux2015binaryconnect}. To compute the gradient of the sign function we use the Straight Through Estimator (STE)~\cite{bengio2013estimating}: $d/dx ~ \text{sign}(x) = {\bf 1}_{|x|\le 1}$. During the training we keep the full precision weights and use Stochastic Gradient Descent (SGD) to gradually update them in back-propagation. In the forward-pass, only the quantized weights are used.

During the training we compute quantizers (for both weights and activations) using the online statistics, i.e., the scalars in a $k$-bits scaled binary quantization (\ref{eqn:kbit_representable}) are computed based on the observed values. During the training we also store the running average of these scalars. During inference we use the stored quantized scalars to improve the efficiency. This procedure is similar to the update of the batch normalization parameters in a standard DNN training~\cite{ioffe2015batch}.

%-------------------------------------------------------------------------
%%%%%%%%% Experiments
%-------------------------------------------------------------------------
\section{Experiments} \label{sec:experiments}
We conduct experiments on the ImageNet dataset~\cite{deng2009imagenet} using the ResNet-18 architecture~\cite{he2016deep}. The details of the architecture and training are provided in Appendix \ref{sec:training_details}. 

We conduct three sets of experiments: (1) evaluate quantization error of activations of a pre-trained DNN, (2) evaluate the quantization error based on the classification accuracy of a post-training quantized network, and (3) evaluate the classification accuracy of during-training quantized networks.
We report the quantization errors of the proposed binary quantization algorithms (least squares 1-bit, 2-bits, ternary, and the greedy foldable quantizations) and compare with the state-of-the-art algorithms BWN-Net~\cite{rastegari2016xnor}, XNor-Net~\cite{rastegari2016xnor}, TWN-Net~\cite{li2016ternary}, DoReFa-Net~\cite{zhou2016dorefa}, ABC-Net~\cite{lin2017towards}, and LQ-Net~\cite{zhang2018lq}.

%-------------------------------------------------------------------------
\subsection{Quantization error of activations}\label{sec:sqr}
To quantify the errors of the introduced binary quantization algorithms we adopt the analysis performed by \cite{anderson2017high}. They show that the angle between $\vx$ and $\vx^q$ can be used as a measure of accuracy of a quantization scheme. They prove that when $\vx^q = \text{sign}(\vx)$ and elements of $\vx$ are i.i.d. $\sim \mathcal{N}(0,1)$, $\angle{(\vx, \vx^q)}$ converges to $\sim 37^{\circ}$ for large $N$.

\begin{figure}[htb!]
\begin{center}
  \includegraphics[scale=0.28]{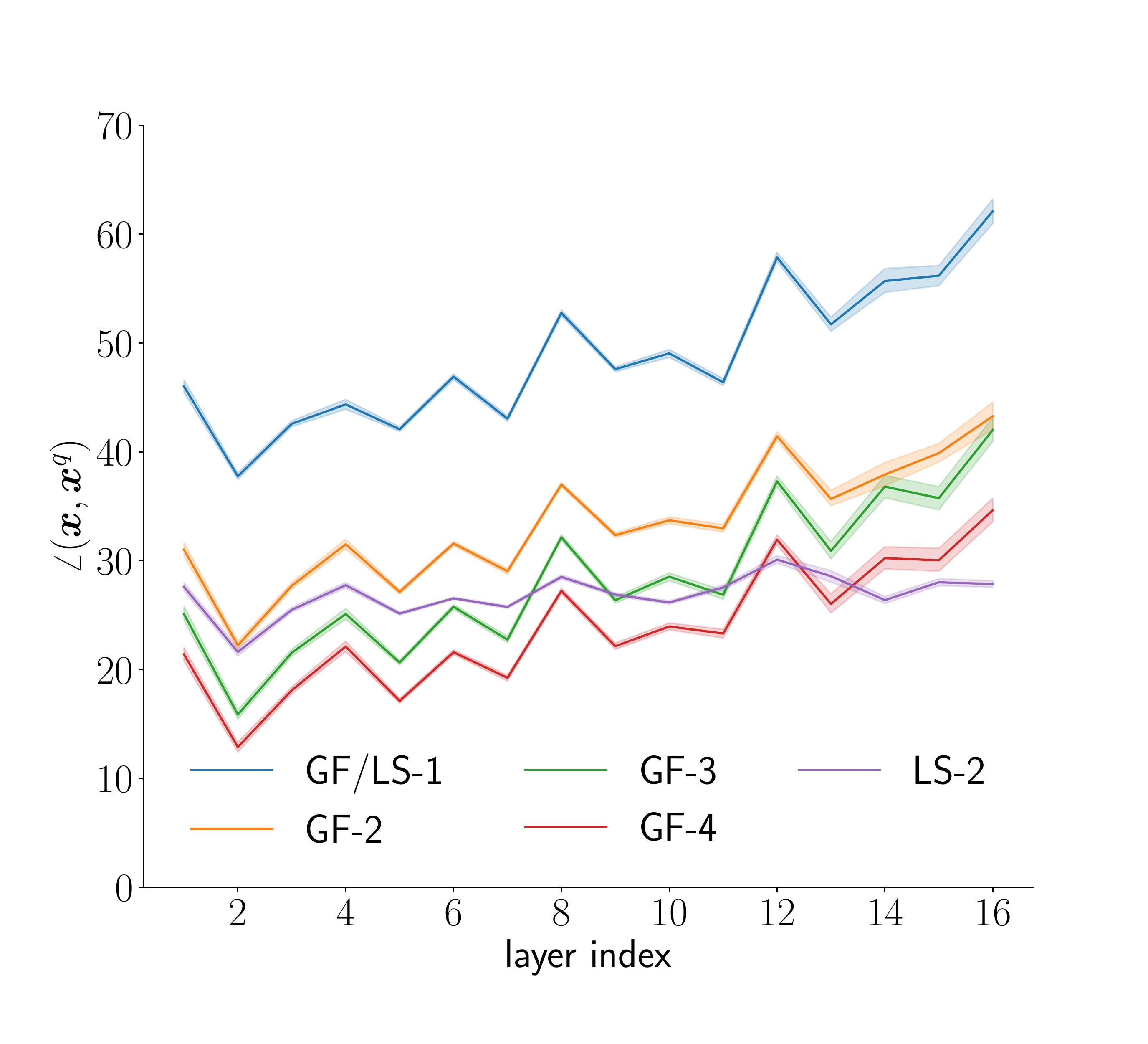}
 \caption{The angle between the full precision and the quantized activations for different layers of a trained full precision ResNet-18 architecture on ImageNet. The 95\% confidence interval over different input images is shown. LS and GF refer to Least Squares and Greedy Foldable, respectively.}
 \label{fig:angles}
 \end{center}
\end{figure}

Here we use the real data distribution.
We trained a full precision network, and computed the activation tensors at each layer for a set of 128 images. In Figure \ref{fig:angles} we show the angle between the full precision and quantized activations for different layers. When the least squares quantization is used, a significant reduction in the angle is observed compared to the greedy algorithm. The least squares 2-bits quantization is even better than the greedy 4-bits quantization for later layers of the network, for which activation tensors have more skewed distribution, make it harder for quantization in form of (\ref{eqn:kbit_representable}). Furthermore, the accuracy of the least squares quantization has less variance with respect to different input images and different layers of the network.

\begin{table}
\begin{center}
 \begin{tabular}{lccccc} \hline
 Method & $k^a$ & $k^w$& Top-1& Top-5\\
 \hline
 \hline
 Post Greedy Foldable & 32 & 1 & 0.1 & 0.5 \\
 Post Greedy Foldable & 32 & 2 & 0.3 & 1.1 \\
 Post Greedy Foldable & 32 & 3 & 1.4 & 4.6 \\
 Post Greedy Foldable & 32 & 4 & 5.3 & 14.1 \\
 Post Least Squares & 32 & 2 & 5.3 & 13.9 \\
  \end{tabular}
 \caption{Validation accuracy of a quantized ResNet-18 trained on ImageNet.
	  $k^a$ and $k^w$ are number of bits to quantize activations and weights, respectively.}
  \label{tab:compare_in}
  \end{center}
\end{table}

%-------------------------------------------------------------------------
\subsection{Post-training quantization}
In this section we apply post-training quantization to the weights of a pre-trained full precision network.
We then use the quantized network for inference and report the classification accuracy.
This procedure can result in an acceptable accuracy for a moderate number of bits (e.g., 16 or 8). 
However, the error significantly grows with a lower number of bits, which is the case in this experiment.
Therefore, we only care about the relative differences between different quantization strategies.
This experiment demonstrates the effect of quantization errors on the accuracy of the quantized DNNs.
The results are shown in Table \ref{tab:compare_in}. When the least squares 2-bits quantization is used, significant accuracy improvement (more than one order of magnitude) is observed compared to the greedy 2-bits quantization, which illustrate the effectiveness of the optimal quantization.

%-------------------------------------------------------------------------
\subsection{During-training quantization}
To achieve higher accuracy we apply quantization during the training, so that the model can adapt to the quantized weights and activations. In Table \ref{tab:compare_ex} we report results from the related works in which ResNet-18 architecture with quantized weights and/or activations is trained on the ImageNet dataset for the classification task.
The proposed least-squares quantization algorithms improve the classification accuracies when compared with the state-of-the-art (with even more number of bits) significantly.
For all of our results we use some of the suggested training setups discussed in the literature (see Appendix \ref{sec:training_details}) to train the BNNs. These changes improved the performance of the baseline 1-bit XNor-Net~\cite{rastegari2016xnor} from 51.2$\%$ to 59.0$\%$. With $2$-bits the least squares quantization algorithm achieves a significantly better accuracy compared to the $2$-bits greedy and $1$-bit, with an identical training setup.

\begin{table}
\begin{center}
  \begin{tabular}{lllll}
  \hline
 Method & $k^a$ & $k^w$& top-1& top-5\\
  \hline
  \hline
   XNor-Net~\cite{rastegari2016xnor} & 1 & 1 & 51.2 & 73.2 \\
   Bi-Real Net~\cite{liu2018bi}& 1 & 1 & 56.4 & 79.5 \\
   XNor-Net++~\cite{bulat2019xnor}&1&1&57.1&79.9\\
  Least Squares (ours) & 1 & 1 & \bf 58.9 &\bf  \bf 81.4 \\
%  Least Squares $2\times$ longer   & 1 & 1 & \bf 60.0 &\bf 81.9 \\
   \hline
   Least Squares (ours)&  T &  1 &\bf 62.0 &\bf 83.6 \\
   \hline
     HWGQ-Net~\cite{cai2017deep}& 2 & 1 &59.6 & 82.2\\
  LQ-Net~\cite{zhang2018lq}& 2 & 1 & 62.6 & 84.3\\
    Greedy Foldable (ours)&  2 &  1 & 62.6 &  84.0 \\
  Least Squares (ours)&  2 &  1 & \bf 63.4&\bf 84.6 \\
   \hline
   DoReFa-Net~\cite{zhou2016dorefa} & 4 & 1 & 59.2 & 81.5\\
   \hline
      SYQ~\cite{faraone2018syq} & 8 & 1 & 62.9 & 84.6\\
      \hline
    DoReFa-Net~\cite{zhou2016dorefa} & 2 & 2 & 62.6 & 84.4\\
    \hline
  ABC-Net~\cite{lin2017towards} & 3 & 3 & 61.0 & 83.2\\
     \hline
   BWN-Net~\cite{rastegari2016xnor} &  32 & 1& 60.8 & 83.0 \\
  Least Squares (ours)&   32 &  1 & \bf 66.1&\bf 86.5\\
   \hline
  TWN-Net~\cite{li2016ternary} & 32 & T & 61.8 & 84.2\\
   \hline
  Full precision baseline &   32 &  32&  69.6 &  89.2 \\
  \end{tabular}
   \caption{Validation accuracy of ResNet-18 architecture on the ImageNet dataset.
   T refers to ternary quantization.
    }
  \label{tab:compare_ex}
\end{center}
\end{table}

%-------------------------------------------------------------------------
%%%%%%%%% CONCLUSION
%-------------------------------------------------------------------------
\section{Conclusion}
In this work, we analyze the accuracy of binary quantization to train DNNs with quantized weights and activations. We discuss methods to improve the accuracy of quantization, namely scaling and using more bits. 

We introduce the rank-$1$ binary quantization, as a general scaling scheme. Based on a singular value analysis we motivate using the scaled binary quantization, a computationally efficient scaling strategy. We define a general $k$-bits scaled binary quantization. We provide provably least squares 1-bit, 2-bits, and ternary quantizations. In addition, we propose a greedy $k$-bits quantization algorithm. We show results for post and during-training quantization, and demonstrate significant improvement in accuracy when least squares quantization is used. We compare the proposed quantization algorithms with state-of-the-art BNNs on the ImageNet dataset and show improved classification accuracies.

\appendix
%-------------------------------------------------------------------------
%%%%%%%%% APPENDIX: OPTIMAL RANK-1
%-------------------------------------------------------------------------
\section{Optimal rank-$1$ binary quantization} \label{sec:opt_rank1}
In this section, we find the optimal rank-1 binary quantization of an $m$ by $n$ matrix $\mX$ discussed in Section \ref{sec:lr_sign_appx}:
\begin{equation}\label{eqn:2bits_opt_problem}
\begin{aligned}
\underset{\mX_1, \mS}{\text{minimize}} \quad &
\| \mX - \mX_1 \odot \mS \|_F\\
\text{s.t.} \quad & \mS \in \{-1,1\}^{m \times n}\\
\quad & \mX_1 \in \R^{m \times n}\\
\quad & \text{rank}(\mX_1) = 1
\end{aligned}
\end{equation}
First, observe that the element-wise multiplication by $-1$ and $+1$ does not change the Frobenius norm. Therefore:
\begin{equation}
\label{eqn:rank1opt}
\begin{aligned}
\min_{\mS, \mX_1} \| \mX - \mX_1 \odot \mS \|_F &= \min_{\mS, \mX_1}  \| (\mX - \mX_1 \odot \mS)\odot \mS \|_F\\ &= \min_{\mS, \mX_1}  \| \mX \odot \mS- \mX_1 \|_F
\end{aligned}
\end{equation}
Furthermore, note that 
\begin{equation}\label{eqn:sumsigma2}
\min_{\mS, \mX_1}  \| \mX \odot \mS- \mX_1 \|_F^2 = \sigma_2^2 (\mX \odot \mS) + \ldots + \sigma_r^2 (\mX \odot \mS)
\end{equation}
Here $\sigma_i (\mX \odot \mS)$ is the $i$'th singular value of $\mX \odot \mS$ and $r$ is its rank. In addition for any $\mS$:
\begin{equation}
\sum_{i=1}^r \sigma_i^2(\mX \odot \mS) = \|\mX \odot \mS\|_F^2 = \| \mX \|_F^2
\end{equation}
Hence, to minimize the sum in (\ref{eqn:sumsigma2}) we need to find an $\mS$ for which $\sigma_1^2(\mX \odot \mS)$ is maximized:
\begin{equation}
\min_{\mS, \mX_1}  \| \mX \odot \mS- \mX_1 \|_F^2 = \|\mX\|_F^2 - \max_{\mS}  \sigma_1^2(\mX \odot \mS)
\end{equation}
$\sigma_1(\mX \odot \mS) =  \| \mX \odot \mS \|_2$ is the 2-norm of $\mX \odot \mS$. Therefore:
\begin{equation}
\max_{\mS}  \sigma_1^2(\mX \odot \mS) = \max_{\mS} \max_{\|\vr\|_2=1}  \| (\mX \odot \mS)\vr \|_2^2
\end{equation}
For any $\mS$ and $\vr \in \R^n$ we have $ \| (\mX \odot \mS)\vr \|_2^2 \le   \| ~\! |\mX| ~\! |\vr| ~\! \|_2^2$ since for $1\le i \le m$ we have $|\sum_j S_{i,j}X_{i,j} r_j| \le \sum_j |X_{i,j}|~\! |r_j|$. Here $| \mX | = \mX \odot ~\! \text{sign}(\mX)$ is the element-wise absolute value of $\mX$. Note that for $\mS = \text{sign}(\mX)$ and $\vr$ with positive values the inequality becomes an equality. Therefore:
\begin{equation}
\max_{\mS} \max_{\|\vr\|_2=1}  \| (\mX \odot \mS)\vr \|_2^2 = \max_{\|\vr\|_2=1}  \| ~\! |\mX| ~\! |\vr| ~\! \|_2^2 
\end{equation}
Observe that the element-wise absolute value does not change the vector norm, i.e. $\|~\! |\vr| ~\! \|_2 = \|\vr\|_2$, and hence $| \vr |$ is a unit vector when $\vr$ is. Also for any $\vr$ we have $  \| ~\! |\mX| ~\! \vr ~\! \|_2^2 \le  \| ~\! |\mX| ~\! |\vr| ~\! \|_2^2$ since for $1\le i \le m$ we have $|~\!\sum_j |X_{i,j}|r_j~\!| \le \sum_j |X_{i,j}| |r_j|$. So we have
\begin{equation} \label{eqn:trunc}
\max_{\|\vr\|_2=1}  \| ~\! |\mX| ~\! |\vr| ~\! \|_2^2 =  \max_{\|\vr\|_2=1}  \| ~\! |\mX| ~\! \vr ~\! \|_2^2 = \sigma_1^2( | \mX |)
\end{equation}
Therefore, we showed that $\mS = \text{sign}(\mX)$ and $X_1$ equal to the best rank-1 approximation of $|\mX|$ (i.e. the first term in its SVD) is a solution of (\ref{eqn:2bits_opt_problem}).$\quad\quad \qed$

\begin{figure}[htb!]
\begin{center}
\includegraphics[scale=0.2]{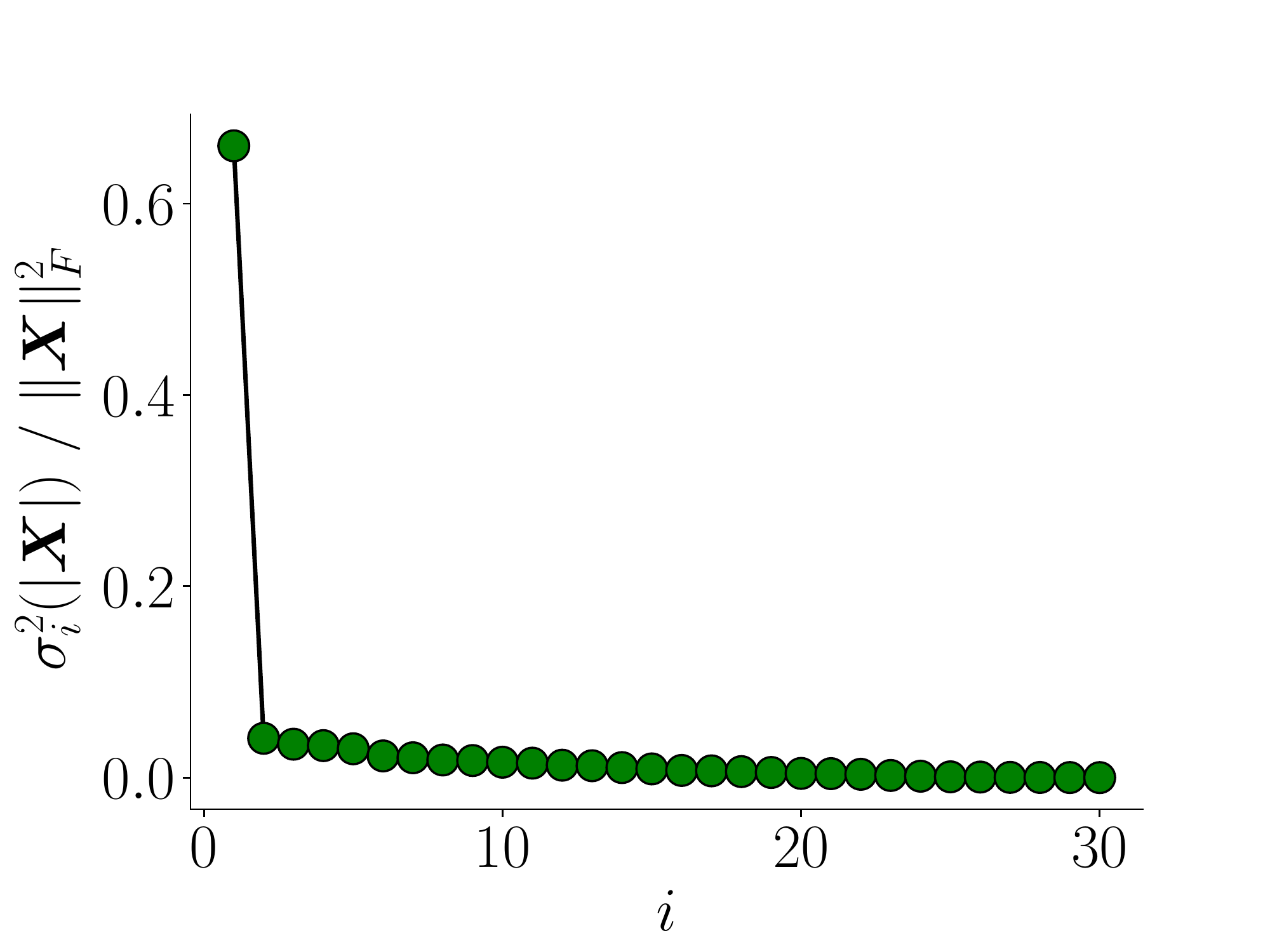}
\includegraphics[scale=0.2]{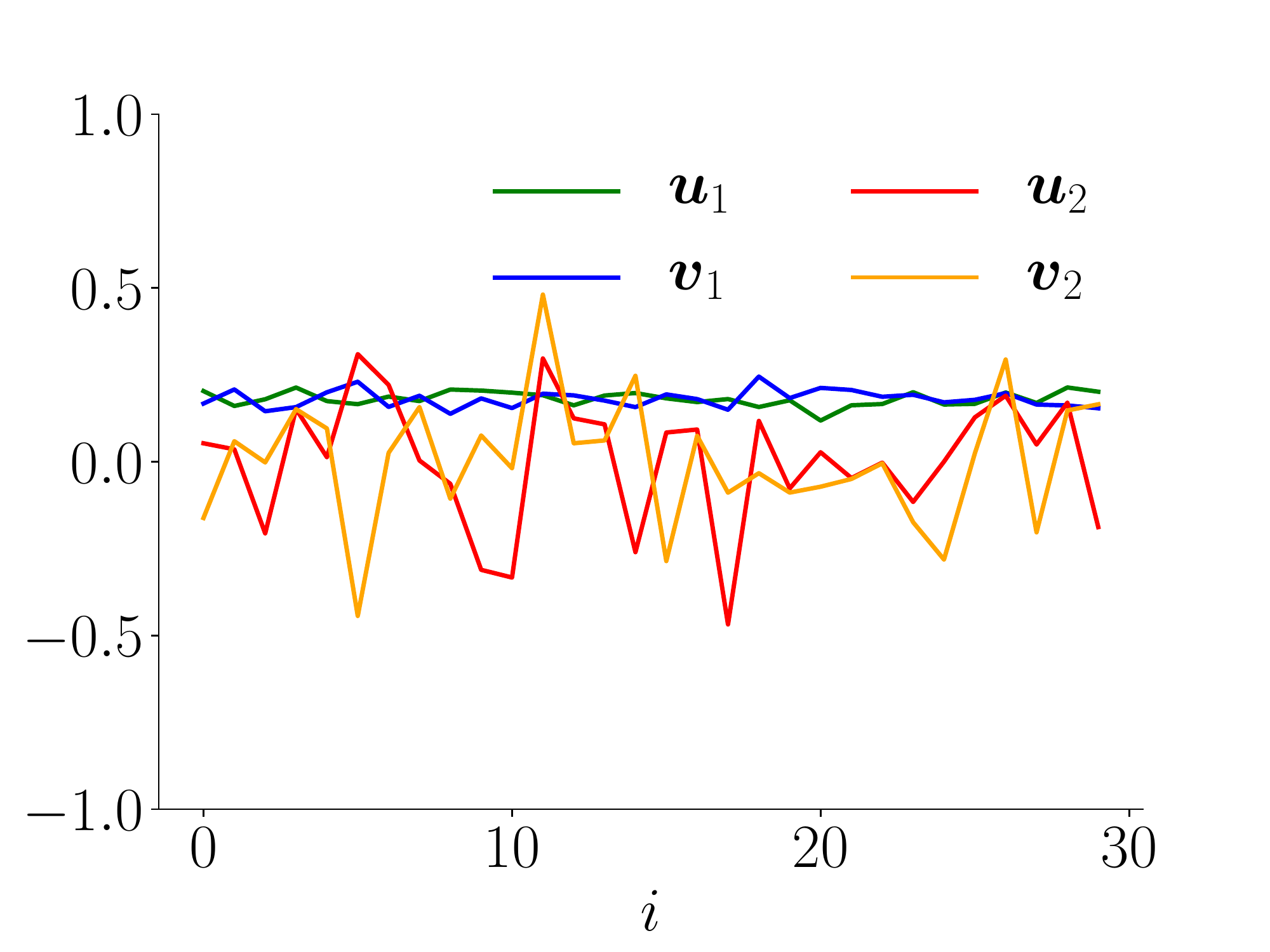}
\end{center}
\caption{{\bf Left}: Distribution of energy for $|\mX| $, where $\mX \in \R^{30 \times 30}$ is a standard normal random matrix. {\bf Right}: Entries of the first left and right singular vectors of $|\mX|$ (shown in green and blue) are almost constant.}
\label{fig:svd_normal}
\end{figure}

Now consider the case that entries of $\mX$ are i.i.d. $\sim \mathcal{N}(0,1)$. In~\cite{bryc2019singular,silverstein1994spectral} the authors show that the largest singular value of a random matrix with i.i.d. entries from a distribution with mean $\mu$ and bounded 4th moment (which is the case for standard folded normal distribution) asymptotes to $\sqrt{mn}\mu$ as $m$ and $n$ are increased (with $m/n \to$constant). Note that $\E[\frac{{\bf 1}^T}{\sqrt{m}} |\mX|~\! |\mX|^T \frac{{\bf 1}}{\sqrt{m}} ]  = \E[\frac{{\bf 1}^T}{\sqrt{n}} |\mX|^T ~\! |\mX| \frac{{\bf 1}}{\sqrt{n}} ] = mn\mu^2$ (with convergence given by the central limit theorem), and therefore, for large matrices the first left and right singular vectors are expected to be almost constant, that is: truncated-SVD-1($|\mX|$) $\approx \mu {\bf 1} {\bf 1}^T$ where $\mu = \sqrt{2/\pi}$. This is shown empirically in Figure \ref{fig:svd_normal}. Therefore, the optimal rank-1 binary quantization captures $2/\pi \simeq 0.64$ of the total energy of $\mX$, making it a good approximation, and can be written as a scalar times a binary matrix.

%-------------------------------------------------------------------------
%%%%%%%%% APPENDIX: OPTIMAL 1-BIT
%-------------------------------------------------------------------------
\section{Least squares $1$-bit binary quantization} \label{sec:opt_scalar_appx}
In this section, we solve (\ref{eqn:order1_sign_appx}). First, observe that:
\begin{equation}
\forall x \in \R: (v-x)^2 < (-v-x)^2 \quad \text{iff} \quad x>0
\end{equation}
Therefore, the optimal choice for function $s$ in (\ref{eqn:order1_sign_appx}) is $s(x)=\text{sign}(x)$. We can rewrite (\ref{eqn:order1_sign_appx}) as follows:
\begin{equation}\label{eqn:opt_scalar_appx2}
\begin{aligned}
\underset{v}{\text{minimize}} \quad & \int_{-\infty}^{+\infty} p(x)(v - |x|)^2 dx\\
\text{s.t.} \quad & v \in \R_{\ge 0}
\end{aligned}
\end{equation}
Setting the gradient of the objective function in (\ref{eqn:opt_scalar_appx2}) with respect to $v$ to zero, we get:
\begin{equation}
v = \frac{ \int_{-\infty}^{+\infty} p(x) |x| dx }{\int_{-\infty}^{+\infty} p(x) dx} = \E_{\rx\sim p} [ |\rx| ] \quad\quad\qed
\end{equation}

%-------------------------------------------------------------------------
%%%%%%%%% APPENDIX: OPTIMAL 2-BITS
%-------------------------------------------------------------------------
\section{Least squares $2$-bits binary quantization} \label{sec:optimal_2bits}
In this section, we solve the following optimization problem corresponding to the least squares 2-bits binary quantization as discussed in Section \ref{sec:2bits_sign_appx}:
\begin{equation}\label{eqn:optimal_2bits_sign_appx}
\begin{aligned}
\underset{v_1,v_2, s_1,s_2}{\text{minimize}} \quad & \int_{-\infty}^{+\infty} p(x)\left(v_1 s_1(x) + v_2 s_2(x)  - x\right)^2 dx\\
\text{s.t.} \quad &s_1, s_2: \R \to \{-1,1\}\\
\quad & v_1 \ge v_2\ge 0
\end{aligned}
\end{equation}
First, we show that the optimal 2-bits binary quantization is foldable, i.e., $\forall x\in \R~ s_1(x) = \text{sign}(x)$ and $s_2(x) = \text{sign}(x - v_1s_1(x))$. Observe that
\begin{equation} \label{eqn:proof_2bits_foldable}
\begin{aligned}
f(x) &= \left(v_1 s_1(x) + v_2 s_2(x)  - x\right)^2\\ &=
~ v_1^2 \left(1 + \frac{v_2}{v_1} s_1(x) s_2(x)  - \frac{s_1(x)x}{v_1}\right)^2\\ &\ge
~ v_1^2 \left(1 + \frac{v_2}{v_1} s_1(x) s_2(x)  - \frac{|s_1(x)x|}{v_1}\right)^2  = g(x)
\end{aligned}
\end{equation}
The inequality in (\ref{eqn:proof_2bits_foldable}) holds because $v_1 \ge v_2$, and therefore, $1 + \frac{v_2}{v_1} s_1(x) s_2(x) \ge 0$. The objective function in (\ref{eqn:optimal_2bits_sign_appx}) is a weighted average of $f(x)$ with non-negative weights. For $x \in \R$ the inequality is strict if $s_1(x) \ne \text{sign}(x)$. In that case, flipping the value of both $s_1(x)$ and $s_2(x)$ reduces $f(x)$ to a strictly smaller value $g(x)$. Hence, the optimal solution of (\ref{eqn:optimal_2bits_sign_appx}) should satisfy $s_1(x) = \text{sign}(x)$ for all $x \in \R$.

For any $v_1$ and $s_1$ if we consider $y = x - v_1 s_1(x)$, the problem reduces to the 1-bit binary quantization for $y$. Based on the result showed in Appendix \ref{sec:opt_scalar_appx} for the optimal solution we have $s_2(x) = \text{sign}(y) = \text{sign}(x - v_1 s_1(x))$. This completes the proof to show that the optimal 2-bits binary quantization is foldable.

Next, we find the optimal values for $v_1$ and $v_2$. Substitute $s_1(x) = \text{sign}(x)$ and $s_2(x) = \text{sign}(x - v_1 s_1(x))$ in (\ref{eqn:optimal_2bits_sign_appx}):
\begin{equation} \label{eqn:opt_2bits_v1v2}
\begin{aligned}
e(v_1, v_2) &= \int_{-\infty}^{+\infty} p(x)\left(v_1 s_1(x) + v_2 s_2(x)  - x\right)^2 dx\\ 
&=\int_{0}^{v_1} q(x) (x - v_1 + v_2)^2 dx\\
&~~+ \int_{ v_1}^{+ \infty} q(x) (x - v_1 - v_2)^2 dx
\end{aligned}
\end{equation}
Here $e(v_1, v_2)$ is the error as a function of $v_1$ and $v_2$, and $q(x) = p(-x) + p(x)$ is the folded distribution function. Assuming the optimal point occurs in the interior of the domain, it should satisfy the zero gradient condition: $\partial e / \partial v_1 = \partial e / \partial v_2 = 0$. Taking derivative from (\ref{eqn:opt_2bits_v1v2}) with respect to $v_1$ and $v_2$ and set it to zero we get:
\begin{equation}\label{eqn:optv1v2_zero_grad}
 \begin{aligned}
v_1 &= \int_{0}^{v_1} xq(x)dx ~+ \int_{v_1}^{+\infty} xq(x)dx\\
~~&+ v_2 \left( \int_{0}^{v_1} q(x) dx - \int_{v_1}^{+\infty} q(x)dx \right) \\
v_2 &= -\int_{0}^{v_1} xq(x)dx + \int_{v_1}^{+\infty}xq(x)dx\\
~~&+ v_1\left(\int_{0}^{v_1} q(x)dx  - \int_{v_1}^{+\infty} q(x)dx  \right)
 \end{aligned}
\end{equation}
Simplifying (\ref{eqn:optv1v2_zero_grad}) results in (\ref{eqn:v1}) and (\ref{eqn:v2}). $\quad\quad\qed$

%-------------------------------------------------------------------------
%%%%%%%%% APPENDIX: TRAINING SETUP DETAILS
%-------------------------------------------------------------------------
\section{Details of training ResNet on ImageNet} \label{sec:training_details}
In this section, we explain the details of how the DNN results reported in this paper are produced using some of the stable good practices from the literature. We use the standard training and validation splits of the ImageNet dataset, without hyper parameters search. We followed a similar architecture as XNor-Net~\cite{rastegari2016xnor}. The convolutional block that we use is depicted in Figure \ref{fig:arch}. We use PReLU~\cite{he2015delving,bulat2019improved} non-linearity before the batch normalization as suggested by~\cite{rastegari2016xnor}. Also, we find it important to use bounded dynamic range, and therefore clip the values to $[-d, d]$.
Without tuning, we picked $d=$ 2, 3, 5, and 8 for $k=$ 1, 2/T, 3, and 4 bits quantizations, respectively. Similar to the other BNNs for the first and last layers we use full precision. Also, as suggested by \cite{liu2018bi} we use full precision {double short-cuts} in ResNet architecture, which adds a small computational/memory overhead. We quantize weights per filter and activations per layer.
%This has about 1\% improvement in top-1 accuracy.
As \cite{cai2017deep} we use first-order polynomial learning-rate annealing schedule (from $2.0e-4$ to $2.0e-7$) and train for 240 epochs with Adam optimizer. When use 120 epochs, with identical setup, we get top-1 accuracies for different activations quantizations (with ls-1 binary weights) as follows: ls-1: 58.1, T:61.1, gf-2:61.1, ls-2:62.4, and fp:65.2. We use smooth labels from a pre-trained full precision teacher \cite{hinton2015distilling} with unit temperature. Quantized networks are initialized randomly.
%Using this learning-rate compared to steps resulted in 1-2 \% accuracy improvement. 
We do not use weight decay. For the data augmentation we use the standard methods used to train full precision ResNet architecture. For training we apply random resize and crop to 224$\times$224, followed by random horizontal flipping, color jittering, and lightening. For test we resize the images to 256$\times$256 followed by a center cropping to 224$\times$224.

\newpage
{\small
\bibliographystyle{ieee_fullname}
\bibliography{nc_quant}
}
\end{document}